\documentclass{article}

 \usepackage[dblblindworkshop, final]{neurips_2026}
\usepackage[utf8]{inputenc} % allow utf-8 input
\usepackage[T1]{fontenc}    % use 8-bit T1 fonts
\usepackage{hyperref}       % hyperlinks
\usepackage{url}            % simple URL typesetting
\usepackage{booktabs}       % professional-quality tables
\usepackage{amsfonts}       % blackboard math symbols
\usepackage{amsmath}
\usepackage{nicefrac}       % compact symbols for 1/2, etc.
\usepackage{microtype}      % microtypography
\usepackage{xcolor}         % colors
\usepackage{booktabs}
\usepackage{tabularx}
\usepackage{tabularray}
\usepackage{multirow}
\usepackage{multicol}
\usepackage{tikz}
\usepackage{pgfplots}
\pgfplotsset{compat=1.18}
\usepackage{caption}
\usepackage{enumitem}
\definecolor{GoogleBlue}{HTML}{4285F4}
\definecolor{GoogleRed}{HTML}{EA4335}
\definecolor{GoogleYellow}{HTML}{FBBC05}
\definecolor{GoogleGreen}{HTML}{34A853}

\newcommand{\bttf}{\textsc{BTTF}}

\workshoptitle{AI for Chip Design}
\title{Back to the Future: Rethinking EDA Infrastructure\\for Agentic Systems in Chip Design Verification}

\author{%
  Je Yang\thanks{Work done while interning at Google.} \\
  Department of Computer Science\\
  Columbia University\\
  New York, NY, USA \\
  \texttt{je.yang@cs.columbia.edu} \\
  \And
  Ivan Lobov \\
  Google DeepMind \\
  London, United Kingdom \\
  \texttt{lobov@google.com} \\
  \And
  Thomas Karpati \\
  Google \\
  Mountain View, CA, USA \\
  \texttt{tkarpati@google.com} \\
}

\begin{document}

\maketitle

\begin{abstract}
The unprecedented computational scale of modern artificial intelligence depends on complex, multi-billion-transistor Systems-on-Chip---yet the workflows that verify these chips remain stubbornly manual. Although Large Language Models (LLMs) have made rapid inroads into Electronic Design Automation (EDA), approximately 74.6\% of existing studies target static Register-Transfer Level (RTL) code generation, leaving post-simulation verification and interactive waveform debugging largely untouched. We introduce \textbf{Back-to-the-Future (\bttf)}, an end-to-end agentic framework that closes this infrastructural gap. \bttf{} distills massive, unstructured simulation dumps into a normalized relational SQLite database and couples it with a collaborative multi-agent orchestration engine that translates natural-language verification queries into schema-aware SQL while correlating signal anomalies with versioned RTL repositories. Across a 150-query benchmark, \bttf{} attains 95.33\% execution accuracy, charting a practical path toward autonomous EDA verification.
\end{abstract}

\vspace{-0.8em}
\section{Introduction}
\vspace{-0.2em}

% Why chip design
Sustaining the computational demands of modern AI beyond the end of Moore's law increasingly hinges on highly heterogeneous Systems-on-Chip (SoCs)~\cite{11503293}. As architectural complexity grows, design flows built on Hardware Description Languages (HDLs) become ever more labor-intensive, iterative, and error-prone~\cite{hu2025uvllm}. Verification alone now consumes over 70\% of the development cycle becoming the primary bottleneck of modern chip design~\cite{cadence2024soc}, and functional bugs that slip into full-chip regression or post-silicon testing carry exponential financial and schedule penalties. Hence, automating register-transfer level (RTL) verification and analysis is an urgent, high-impact goal for the Electronic Design Automation (EDA) community.

% AI for chip design
The rapid advances of Large Language Models (LLMs) has naturally inspired a wave of AI-assisted EDA research~\cite{kang2025fveval, ping2026verimoa}. This wave, however, is strikingly lopsided. While approximately 74.6\% of LLM-for-hardware studies concentrate on front-end RTL code generation and synthesis~\cite{abdollahi2024hardware}, post-simulation waveform debugging and dynamic verification remain largely underexplored. Since hardware development demands rigorous testing as much as design synthesis, this neglect leaves a substantial share of the automation opportunity untapped.

Even the few LLM-aided verification workflows that do exist operate almost exclusively on static source code or toy-scale simulation logs, whereas interactive post-simulation debugging is a fundamentally harder problem. When a dynamic simulation fails, engineers manually scrub through terabytes of temporal dumps in graphical waveform viewers, tracing illegal transitions back to their origin in the HDL hierarchy. Frontier LLM agents are locked out of this loop by what we term \emph{Infrastructural Impedance}. Raw, ASCII-based waveform dumps exceed model context windows by orders of magnitude, carry no structured semantic index, and expose no native, low-latency query interface that an autonomous agent could invoke.

Here, we propose \textbf{Back-to-the-Future (\bttf)}, an end-to-end agentic framework that enables natural-language querying for root-cause analysis and formal verification over complex RTL waveforms. Rather than forcing waveforms into an LLM's context, \bttf{} revisits a classical EDA paradigm. It recasts unstructured temporal waveforms as a normalized relational SQLite database, then pairs this schema with a collaborative multi-agent engine that plans, queries, and audits on the engineer's behalf. On 150 expert-annotated queries, \bttf{} reaches 95.33\% overall accuracy, and its multi-agent design outscores a monolithic single-agent baseline by 17.3\% on a five-criteria quality audit.
\vspace{-0.3em}
\section{Related Work}
\vspace{-0.3em}

\textbf{LLM-Driven Hardware Verification.}
LLMs have demonstrated remarkable capacity across EDA research, assisting developers with bug localization, code generation, and automated verification~\cite{hu2025uvllm, kang2025fveval, ping2026verimoa, zhao2025mage, mastropaolo2023robustness, alvanaki2026notb}. For complex verification tasks, state-of-the-art systems increasingly adopt multi-agent orchestration flows that split hardware synthesis into specialized roles for generation, testbench simulation, and formal counterexample refinement~\cite{ping2026verimoa, zhao2025mage, pan2026kvflow}. Yet these frameworks operate predominantly on static source code, compile-time assertion lists, or toy-scale simulation logs, implicitly assuming an LLM can localize functional errors from static HDL traces or terse compiler messages alone.

\textbf{Key Limitation and Our Observation.}
The closest attempt to bridge this gap equips a coding agent with an Abstract Syntax Tree (AST)-based waveform tracing tool~\cite{ho2025verilogcoder}, but such traces remain tied to small, code-centric examples rather than production dumps. \bttf{} instead builds on a complementary observation: SQL generation is an emergent capability of LLMs, rooted in their deep semantic mapping from natural language to structured code~\cite{gao2023text}. Once waveforms are made \emph{relational}, decades of database machinery (e.g., indexing, aggregation, low-latency filtering) become directly accessible to language agents.
\vspace{-0.2em}
\section{\bttf{} Framework} \label{sec:framework}
\vspace{-0.1in}

\begin{figure*}[h]
\centering
   \includegraphics[width=0.98\textwidth]{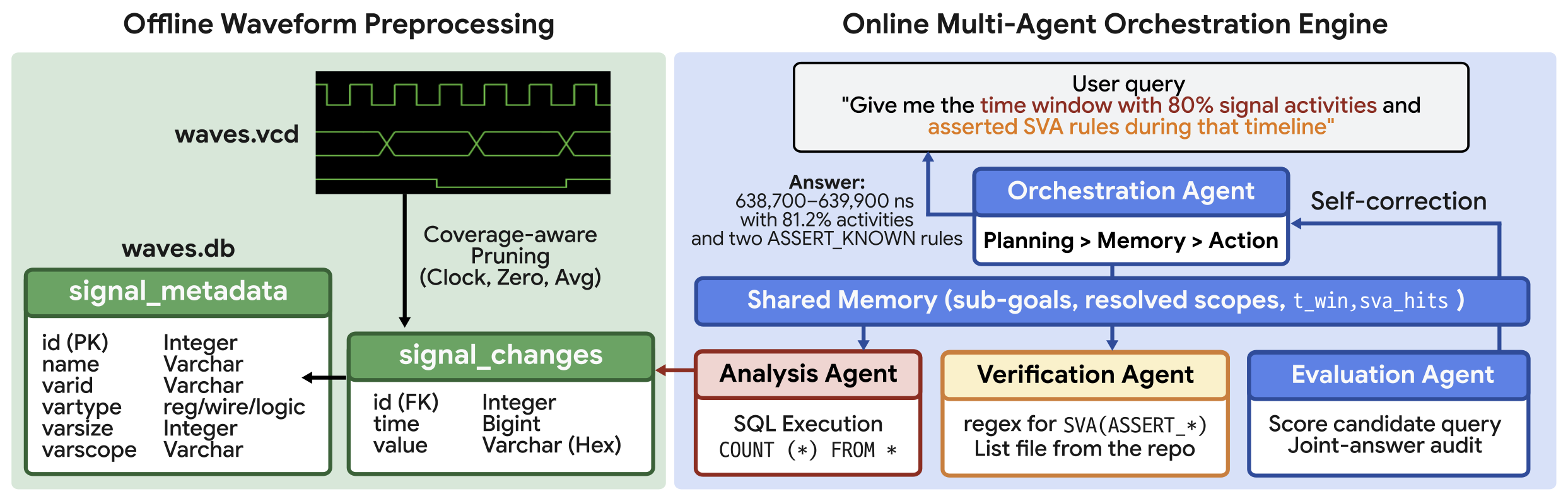}
   \hfil
\caption{Overview of \bttf{} framework.}
\label{overall}
\vspace{-0.1in}
\end{figure*}
  
As Figure~\ref{overall} illustrates, \bttf{} comprises two coupled subsystems on opposite sides of a single relational contract: (1) an offline \textbf{Waveform Preprocessing Pipeline} that compresses bulky simulation dumps into an optimized relational SQLite schema, and (2) an online \textbf{Agentic Orchestration Engine} that translates natural-language verification queries into schema-aware SQL and carries out the verification itself.

\subsection{Waveform Preprocessing Pipeline} Standard EDA simulators emit proprietary binary waveforms (\texttt{.fsdb}) or ASCII Value Change Dump (\texttt{.vcd}) files; for SoC-scale blocks, a raw VCD text file routinely exceeds 10\,GB---far beyond any prospect of direct context embedding. The \bttf{} pipeline therefore extracts the target hierarchies and normalizes them into a relational SQLite database (\texttt{waves.db}) built around two primary tables:
\begin{itemize}[leftmargin=*, nosep]
  \item \texttt{signal\_metadata} captures the static signal declarations (RTL signal name and its variable type, etc.) with id.
  \item \texttt{signal\_changes} records the temporal behavior as timestamped transitions as a \texttt{(time, value)} with a foreign key to refer metadata.
\end{itemize}

Separating \emph{who a signal is} from \emph{what it does} lets agents reason over structure and time independently, joining the two only when a query demands it. A na\"ive translation of every value change would simply move the scalability problem from the filesystem into the database. \bttf{} instead applies coverage-aware pruning during on-the-fly database generation:
\vspace{-0.3em}
\begin{itemize}[leftmargin=*, nosep]
  \item \textbf{Clock pruning}. High-frequency clock toggles dominate table size while carrying little debug information; since every row in \texttt{signal\_changes} is timestamped, clock transitions can be skipped at insertion time without losing temporal anchoring.
  \item \sloppy \textbf{On-the-fly inactive-signal pruning} (\textsc{zero}/\textsc{avg}). To cope with duplicate VCD identifiers and inactive nets, the ingestion engine maintains a lightweight hash map (\texttt{vcd\_id\_to\_db\_id}) alongside a per-signal transition counter (\texttt{signal\_change\_count}). This bookkeeping enables two graduated policies: dropping statically silent zero-toggle wires (\textsc{zero}), or additionally dropping nets whose toggle rate falls below the design-wide average (\textsc{avg}).
  \item \textbf{Batched insertion.} Row-by-row insertion incurs severe SQLite transaction overhead. Grouping transitions into batched writes removes this bottleneck at ingestion time.
\end{itemize}

\vspace{-0.3em}
\subsection{LLM-Driven Agent Orchestration}
 % \emph{(Right)} The online engine decomposes a natural-language query into sub-goals, routes them to specialized agents that exchange results only through a shared memory, and audits the merged answer. Colored spans in the query mark which agent resolves each fragment.
On top of the generated \texttt{waves.db}, \bttf{} conducts verification through a collaborative multi-agent architecture. The main \textbf{Orchestration Agent} decomposes a natural-language query into sub-goals, routes them to specialized agents with own toolsets, and audits the merged answer.
\vspace{-0.3em}
\begin{itemize}[leftmargin=*, nosep]
  \item \textbf{Analysis Agent.} Armed with schema-aware multi-query execution tools, it composes SQLite queries over the two tables to handle scope discovery, event inspection, and toggle counting. For complex tasks such as power hotspot detection, it leverages an iterative function-calling loop to execute multi-turn queries and aggregate multi-cycle signal statistics.
  \item \textbf{Verification Agent.}  Tying waveform anomalies back to version-controlled source by evaluating SVA rules (\texttt{ASSERT}, \texttt{ASSERT\_STATIC}, \texttt{ASSERT\_NEVER}, \texttt{ASSERT\_KNOWN}, \texttt{COVER}) through the regular expression (REGEX) assignment.
  \item \textbf{Evaluation Agent.} Before any root-cause analysis is returned, this agent audits candidate queries against five criteria: (1) \emph{correctness}---filtering, aggregation, and join fidelity to user intent; (2) \emph{completeness}---inclusion of all required columns and time conditions; (3) \emph{efficiency}---proper indexing and avoidance of redundant joins; (4) \emph{schema adherence}---strict conformance to the two-table schema; and (5) \emph{context understanding}---correct inference of implicit simulation timescales and hierarchy scopes.
\end{itemize}
\vspace{-0.3em}
Under the orchestrator, agents communicate solely through a dynamic session context maintained across turns; together with a self-correction loop on failed queries, this shared-memory discipline lets each agent stay narrowly specialized without losing global coherence. Furthermore, when an assertion rule is violated even after self-correction, the agent autonomously isolates the exact violation timestamp and pinpoints the problematic sub-signals responsible for the failure (see Appendix~\ref{apdx:failure} for detailed root-cause diagnostics).
\vspace{-0.3em}
\section{Results}
\vspace{-0.3em}

\textbf{Experimental setup.} All agents are served through the Gemini API with Gemini-2.5-Pro as the default model. Our benchmark spans multiple SoC design hierarchies of varying complexity (4.2\,GB of VCD, 18{,}450 nets), from which we extract simulation windows ranging from narrow 50~ns high-resolution glitch-debug windows to 800~ns full-transaction simulation dumps. We curate 150 natural-language verification queries representative of daily engineering workflows, each paired with an expert-annotated golden answer.

\textbf{Preprocessing Performance.} \begin{figure}[t!]
\centering
% === [Left 55%: Compact Preprocessing Table] ===
% 정렬 옵션을 [c]에서 [b]로 변경
\begin{minipage}[b]{0.6\textwidth}
\centering
\footnotesize
\begin{tblr}{
  width = 1.1\linewidth,
  colspec = {@{} l r r r r @{}},
  column{even} = {r},
  column{3} = {r}, 
  column{5} = {r,font=\bfseries},
  cell{1}{2} = {font=\bfseries},
  cell{1}{3} = {font=\bfseries},
  cell{1}{4} = {font=\bfseries},
  cell{3}{3} = {font=\bfseries},
  cell{4}{3} = {font=\bfseries},
  cell{5}{3} = {font=\bfseries},
  hline{1,6} = {-}{0.08em},
  hline{2} = {-}{0.05em},
}
\textbf{Pruning Strategy} & Footprint & Reduc. & Latency & Coverage \\
\textbf{Fully Unpruned} & 1,420 MB & 0.00\% & 410.5 ms & 100.00\% \\
\textbf{Clock Removed} & 480 MB & 66.20\% & 125.2 ms & 98.51\% \\
\textbf{Zero-Toggle Pruned} & 395 MB & 72.18\% & 94.8 ms & 100.00\% \\
\textbf{Below-Avg Pruned} & 185 MB & 86.97\% & 41.6 ms & 64.23\%
\end{tblr}
\vspace{-0.1in}
\captionof{table}{Trade-off of waveform preprocessing across pruning strategies.}
\vspace{-0.1in}
\label{tab:preprocessing}
\end{minipage}%
\hfill
% === [Right 40%: Clean Batch Insertion Bar Chart] ===
% 너비를 0.4\textwidth에서 0.40\textwidth로 유지하고 정렬 옵션을 [b]로 변경
\begin{minipage}[b]{0.37\textwidth}
\centering
\begin{tikzpicture}
\begin{axis}[
  width=1.1\linewidth,
  height=3.1cm,
  ybar,
  bar width=12pt,
  enlarge x limits=0.18,
  legend style={at={(0.5,1.15)}, anchor=south, legend columns=-1, draw=none},
  ylabel={Time (s)},
  ylabel style={font=\footnotesize},
  yticklabel style={font=\scriptsize},
  symbolic x coords={Batch 1, Batch 10, Batch 50, Batch 100},
  xtick=data,
  xticklabel style={font=\scriptsize, rotate=20, anchor=north east},
  nodes near coords,
  nodes near coords style={font=\scriptsize},
  ymin=0, ymax=280,
  axis lines*=left,
  ymajorgrids=true,
  grid style=dashed,
]
\addplot[fill=GoogleBlue, draw=black] coordinates {
  (Batch 1, 240.5) (Batch 10, 100.2) (Batch 50, 85.4) (Batch 100, 83.2)
};
\end{axis}
\end{tikzpicture}
\vspace{-0.2in}
\captionof{figure}{DB generation time speedup with batch insertion.}
\label{fig:preprocessing}
\vspace{-0.1in}
\end{minipage}
\end{figure} 
We first quantify the cost of turning a waveform into an agent-ready database. As Table~\ref{tab:preprocessing} shows, suppressing clock insertion alone shrinks the footprint by 66.20\% and cuts query latency by $3.28\times$ while preserving 98.51\% of functional toggle coverage. Additionally pruning untoggled signals pushes the reduction to 72.18\%, while retaining full coverage over active nets. Aggressive below-average pruning compresses furthest but sacrifices 35.77\% of coverage, marking the boundary of safe reduction. Relational schema optimization is thus not an implementation detail but a prerequisite for responsive agent interaction. Figure~\ref{fig:preprocessing} confirms ingestion-side scalability. Batched insertion via SQL \texttt{EXECUTEMANY} accelerates database generation by up to $2.89\times$ over single-row insertion, removing the I/O bottleneck for multi-terabyte waveforms.

\textbf{BTTF Framework Performance.} \begin{table}[t]
  \caption{Comprehensive taxonomy of verification tasks, example natural language user queries, and proof-of-concept execution accuracy.}
  \label{tab:online}
  \centering
  \small
  \begin{tabularx}{\textwidth}{l l X l}
    \toprule
    \textbf{Domain} & \textbf{Task / Query Type} & \textbf{Example Natural Language Query} & \textbf{Accuracy} \\
    \midrule
    \textbf{Debugging} & \textbf{Scope/Net Discovery} & ``List all logics in the  \texttt{moduleA}.'' & \textbf{100.00\%} \\
    \addlinespace[0.5em]
     & \textbf{RTL Attribution} & ``Find the RTL file related to \texttt{logicA}.'' & \textbf{87.50\%} \\
    \addlinespace[0.5em]
    & \textbf{Value Inspection} & ``What is the value of \texttt{logicA} at time \texttt{timeA}?'' & \textbf{95.00\%} \\
    \addlinespace[0.5em]
    & \textbf{Value Change Count} & ``How many times has \texttt{logicA} changed?'' & \textbf{100.00\%} \\
    \addlinespace[0.5em]
    & \textbf{Event Timing Check} & ``When is net \texttt{logicA} first set to \texttt{valueA}?'' & \textbf{100.00\%} \\
    \midrule
    \textbf{Analyzing} & \textbf{Toggle Coverage} & ``Calculate overall toggle coverage in \texttt{jobA.db}.'' & \textbf{91.67\%} \\
    \addlinespace[0.5em]
    & \textbf{Power Hotspot} & ``Give me the 100ns window with max signal activities.'' & \textbf{91.67\%} \\
    \midrule
    \textbf{Verification} & \textbf{Static SVA Check} & ``Verify \texttt{ASSERT\_STATIC(...)}.'' & \textbf{84.62\%} \\
    \addlinespace[0.5em]
    & \textbf{Dynamic SVA Check} & ``Run rule \texttt{ASSERT\_KNOWN(...)}.'' & \textbf{90.91\%} \\
    \bottomrule
  \end{tabularx}
  \vspace{-0.1in}
\end{table}

Table~\ref{tab:online} summarizes \bttf{}'s coverage across routine debugging, toggle-based performance analysis, and formal verification. Over the full 150-query taxonomy, \bttf{} achieves 95.33\% overall accuracy---near-perfect on debugging primitives and consistently reliable on the harder analysis and verification tasks. A detailed breakdown of representative verification queries, agent tool invocations, and synthesized output responses across all task categories is provided in Appendix~\ref{apdx:datail} (Table~\ref{tab:appendix_taxonomy_execution}).

We further compare a monolithic single-agent design against our collaborative architecture, with an autorater judging outputs on the Evaluation Agent's five criteria. While a monolithic single agent achieves faster raw response times, it frequently suffers from context clutter, schema hallucination, and degraded causal tracing on multi-hop verification tasks, scoring only 3.35/5.00 on the Evaluation audit. In contrast, \bttf{} cleanly decouples schema-aware SQL generation from RTL source inspection and evaluation, achieving a 17.3\% higher reasoning quality score and near-perfect execution accuracy. This specialized delegation incurs an average 1.96$\times$ latency overhead, representing a practical and necessary trade-off to ensure robust, hallucination-free verification in production EDA workflows.

\vspace{-0.8em}
\section{Discussion and Conclusion}
\vspace{-0.8em}
Our \bttf{} argues, and empirically supports, a simple thesis. The binding constraint on agentic waveform debugging is data infrastructure, not model capability. By recasting terabyte-scale waveforms into a compact, vendor-agnostic relational schema, \bttf{} turns an intractable context problem into a routine text-to-SQL problem---one at which modern LLMs already excel---and its orchestration engine converts that access into grounded, audited root-cause analysis. The 95.33\% benchmark accuracy and 17.3\% quality gain from agent specialization suggest that autonomous EDA verification will be reached through better interfaces between agents and design data rather than ever-larger models alone. 
To scale across larger full-chip designs, streaming ingestion, standardized scope-path normalization, and a waveform-to-RTL symbol index are the natural next steps; evaluating larger frontier reasoning models to tackle multi-hop causal tracing over these stores remains an exciting direction for future work. 
More broadly, the recipe behind \bttf{}---normalize an opaque EDA artifact into a queryable store, then let specialized agents operate over it---extends naturally beyond waveforms to regression logs, coverage databases, and power traces. We therefore view \bttf{} less as a point tool than as a template for rebuilding EDA infrastructure to meet agents halfway, and we hope it encourages the community to treat data interfaces as a first-class research object in AI-driven chip design.

\bibliographystyle{unsrt}
\bibliography{ref}

%%%%%%%%%%%%%%%%%%%%%%%%%%%%%%%%%%%%%%%%%%%%%%%%%%%%%%%%%%%%

\appendix

\section{Appendix: Detailed Query Execution Workflow and Output Analysis}
\subsection{Handling Assertion Failures and Root-Cause Diagnostics} \label{apdx:failure}
In standard simulation logs, assertion checkers typically emit binary pass/fail flags without pinpointing the originating nets. Within \bttf{}, when an assertion rule (such as \texttt{ASSERT\_NEVER} or \texttt{ASSERT\_KNOWN}) fails during dynamic checking, the agent automatically extracts the exact violation timestamp $t_{\text{violation}}$ where the condition evaluates to true or where an invalid \texttt{X}/\texttt{Z} state is detected. It then initiates backward causal tracing by querying the prior values of all antecedent operands in the assertion expression at $t \le t_{\text{violation}}$. Consequently, the final diagnostic response explicitly highlights the specific sub-signal responsible for the failure (e.g., {``Assertion failed at $t=6929\text{ ns}$ due to net \texttt{req\_sel} transitioning to unknown state \texttt{x} while \texttt{clk} was high''}), bridging the gap between high-level assertion monitoring and low-level root-cause debugging.

\begin{table*}[t!]
  \caption{Detailed taxonomy of verification queries, agent tool invocations, and synthesized output responses from benchmark evaluation.}
  \label{tab:appendix_taxonomy_execution}
  \centering
  \scriptsize
  \renewcommand{\arraystretch}{1.2}
  \begin{tabularx}{\textwidth}{l l p{0.28\textwidth} X}
    \toprule
    \textbf{Domain} & \textbf{Task / Query Type} & \textbf{Natural Language User Query} & \textbf{\shortstack[l]{Agent Action with Tool Invocation \\ Generated Output Response}} \\
    \midrule
    \multirow{6}{*}{\textbf{Debugging}} 
    & \textbf{Scope/Net Discovery} 
    & ``List all the nets in \texttt{moduleA}.'' 
    & \textbf{Action:} \texttt{find\_signal\_in\_scope(db, "moduleA")} \newline
      \textbf{Output:} \texttt{["bank\_id", "bank\_addr"]} (Found 2 struct field signals) \\
    \cmidrule(lr){2-4}
    & \textbf{Signal Scope Discovery} 
    & ``What is the scope name of the signal \texttt{port\_id}?'' 
    & \textbf{Action:} \texttt{find\_scope\_by\_signal\_name(db, "port\_id")} \newline
      \textbf{Output:} \texttt{moduleA\_ep\_\{leaf, master, follower\}\_id} \\
    \cmidrule(lr){2-4}
    & \textbf{RTL Attribution} 
    & ``Find the RTL file related to signal \texttt{signalA.cnt}.'' 
    & \textbf{Action:} \texttt{find\_source\_module("soc\_top", ".../moduleA/rtl/", "cnt")} \newline
      \textbf{Output:} \texttt{design/.../moduleA.sv:L142} (\texttt{logic [7:0] cnt}) \\
    \cmidrule(lr){2-4}
    & \textbf{Value Inspection} 
    & ``What is the value of signal \texttt{ctrl\_status} at time 7155316?'' 
    & \textbf{Action:} \texttt{get\_value\_of\_signal\_at\_time(db, "ctrl\_status", 7155316)} \newline
      \textbf{Output:} Value: \texttt{8184} (\texttt{0x1FF8} in hex) at $t = 7,155.316\text{ ns}$ \\
    \cmidrule(lr){2-4}
    & \textbf{Value Change Count} 
    & ``What is the number of toggles of \texttt{credit\_stall}?'' 
    & \textbf{Action:} \texttt{get\_signal\_toggle(db, "credit\_stall", 6800000, 8000000)} \newline
      \textbf{Output:} \texttt{4 toggles} across simulation window $[680\text{ ns}, 800\text{ ns}]$ \\
    \cmidrule(lr){2-4}
    & \textbf{Event Timing Check} 
    & ``When is the first time the signal \texttt{req\_accept[7]} is set to 42?'' 
    & \textbf{Action:} \texttt{get\_first\_time\_of\_signal\_value(db, "req\_accept[7]", "42", 181.5us, 184us)} \newline
      \textbf{Output:} First transition detected at $t = 182,031.57\text{ ns}$ (\texttt{182031570 ps}) \\
    \midrule
    \multirow{2}{*}{\textbf{Analyzing}} 
    & \textbf{Toggle Coverage} 
    & ``What is the toggle coverage of the scope \texttt{pmu\_events}?'' 
    & \textbf{Action:} \texttt{get\_scope\_toggle\_coverage(db, "pmu\_events")} \newline
      \textbf{Output:} \texttt{14.70\%} toggle coverage (21 active nets / 143 total nets in scope) \\
    \cmidrule(lr){2-4}
    & \textbf{Power Hotspot} 
    & ``Give me a time range with the maximum signal activity.'' 
    & \textbf{Action:} Iterative rolling-window query across sliding 100~ns intervals \newline
      \textbf{Output:} Peak activity window: \texttt{[7612 ns, 9944 ns]} with \texttt{27 total signal toggles} \\
    \midrule
    \multirow{2}{*}{\textbf{Verification}} 
    & \textbf{Static SVA Check} 
    & ``Verify rule \texttt{ASSERT\_STATIC(req\_mem\_width, ...)}.'' 
    & \textbf{Action:} \texttt{verify\_static\_rule(rule\_name, macro\_def, rtl\_params)} \newline
      \textbf{Output:} \textbf{PASSED} (\texttt{64 == (48 + \$clog2(65536))} $\rightarrow$ \texttt{64 == 64}, True) \\
    \cmidrule(lr){2-4}
    & \textbf{Dynamic SVA Check} 
    & ``Run the rule \texttt{`ASSERT\_KNOWN(req\_sel\_is\_known\_A, req\_sel, clk, rst)}.'' 
    & \textbf{Action:} \texttt{verify\_known\_rule(db, signal="req\_sel", clk="clk", rst="rst\_n")} \newline
      \textbf{Output:} \textbf{PASSED} (Zero \texttt{X}/\texttt{Z} unknown states observed on \texttt{req\_sel} at clock edges) \\
    \bottomrule
  \end{tabularx}
\vspace{-0.2in}
\end{table*}

\subsection{End-to-End Query-to-Response Pipeline} \label{apdx:datail}
Table~\ref{tab:appendix_taxonomy_execution} illustrates concrete examples of how the \bttf{} framework processes natural language user queries into structured tool executions and engineer-facing diagnostic answers. When an engineer submits a verification question (e.g., ``List all the nets in \texttt{moduleA}''), the Orchestration Agent routes the query to the Analysis Agent. The agent leverages schema context to map the natural language scope into a structured function invocation (\texttt{find\_signal\_in\_scope}) or a relational SQL join across \texttt{signal\_metadata} and \texttt{changes}. By executing deterministic queries over the indexed \texttt{waves.db}, the system retrieves exact timestamps, signal values in hexadecimal or decimal format, and transition counts without context window truncation or string hallucination. Furthermore, when queries span both temporal waveform behavior and design intent (e.g., {``Find the RTL file related to \texttt{signalA.cnt}''}), the agent correlates the resolved hierarchy with the versioned repository to extract exact line numbers and declaration semantics.
% %%%%%%%%%%%%%%%%%%%%%%%%%%%%%%%%%%%%%%%%%%%%%%%%%%%%%%%%%%%%

% \newpage
% \input{checklist.tex}

\end{document}